\documentclass{article}
\usepackage{ijcai26}

\usepackage{times}
\usepackage{soul}
\usepackage{url}
\usepackage[hidelinks]{hyperref}
\usepackage[utf8]{inputenc}
\usepackage[small]{caption}
\usepackage{graphicx}
\usepackage{amsmath}
\usepackage{amsthm}
\usepackage{booktabs}
\usepackage{algorithm}
\usepackage{algorithmic}
\usepackage{tabularx}

\title{A Scalable Cross-Domain Event Extraction System \\via a Unified Generative Training Framework}

\author{
Siting Liang$^{1,2}$
\and
Omar Adjali$^1$\and
Omair Shahzad Bhatti$^1$\And
Daniel Sonntag$^{1,2}$\\
\affiliations
$^1$German Research Center for Artificial Intelligence\\
$^2$Carl von Ossietzky University of Oldenburg\\
\emails
\{siting.liang, omar.adjali, omair\_shahzad.bhatti, daniel.sonntag\}@dfki.de}

\begin{document}
\maketitle

\begin{abstract}
Event extraction is fundamental to information extraction. Prior approaches often separate event detection and argument extraction or depend on dataset-specific designs, limiting scalability and cross-domain generalization. We propose a unified generative sequence-to-sequence framework that performs event extraction subtasks jointly and supports both pipeline and end-to-end configurations. We fine-tune pretrained language models on multiple event datasets across diverse domains, enabling a single model to retain domain-specific semantics while generalizing over large and evolving label spaces. We demonstrate these capabilities through a web-based application tailored for researchers and practitioners. The platform supports document upload, schema-aware event extraction, visualization of triggers and arguments, and comparison of different extraction configurations across domains.
\end{abstract}


\section{Introduction}

\begin{figure}
    \centering
    \includegraphics[width=\linewidth]{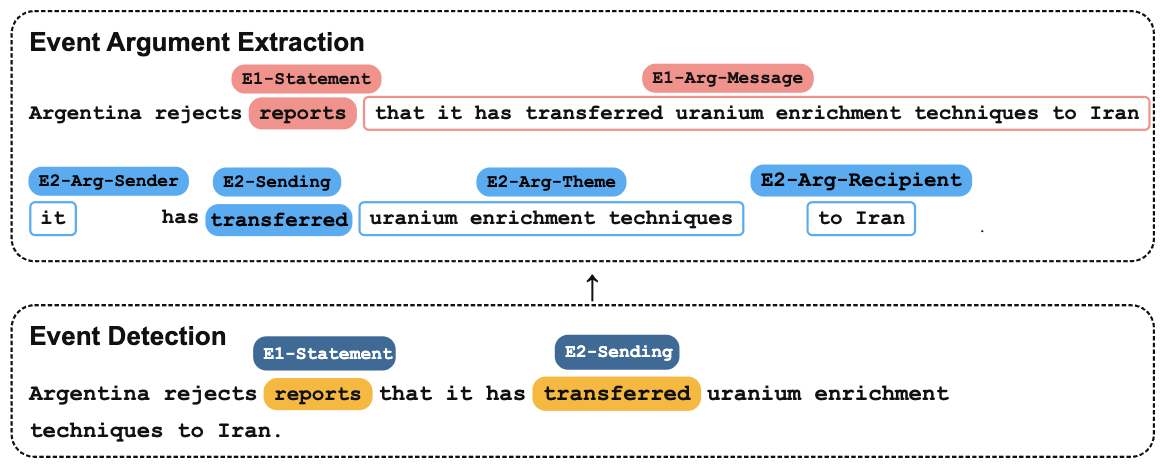}
    \caption{(1) Event Detection identifies event triggers and assigns event types, followed by (2) Event Argument Extraction, which extracts arguments and labels them with semantic roles.}
    \label{fig:example}
\end{figure}

Event extraction (EE) is a fundamental information extraction task that converts unstructured text into structured event representations, which serve as a critical foundation for knowledge base construction~\cite{yang-etal-2019-exploring}. It is typically formulated in two stages: Event Detection (ED), which identifies triggers and classifies their types, and Event Argument Extraction (EAE), which identifies arguments and assigns them semantic roles, as illustrated in Figure~\ref{fig:example}. Many prior methods design dedicated models or learning objectives for individual subtasks, often organized in pipelined architectures~\cite{Zhang21amrie,Hsu23tagprime}. These approaches are often tied to specific annotation schemas and domains, making them difficult to scale or generalize across datasets with different event definitions and label spaces and requiring extensive re-engineering and retraining for new domains. Related cross-domain information extraction studies have shown that pre-trained language models and unified semantic abstractions can support transfer across heterogeneous domain-specific label spaces~\cite{liang-etal-2023-cross}.

More recently, unified information extraction frameworks and large language models have explored general-purpose EE by modeling the task in an end-to-end (E2E), generative manner~\cite{Li21bartgen,Hsu22degree,Hsu23ampere,wang2023instructuie,adjali2026aligning}. However, these methods often use prompts to encode entire event schemas and are typically evaluated on datasets with relatively small label sets~\cite{sainz2024gollie,gui2024iepile}. Furthermore, their high computational requirements make scaling to large or complex event taxonomies impractical. Efforts have also sought to address scalability by constructing large ontology datasets \cite{Li23cedar,rams_ebner2020mult}. These datasets primarily rely on weak supervision signals without domain-specific adaptation, which limits their ability to capture domain-specific context. This underscores the need for a unified EE system that also operates robustly across domains, datasets, and large-scale annotation schemas with minimal redesign demands. 

\begin{figure*}[h!]
    \centering
    \includegraphics[width=\textwidth]{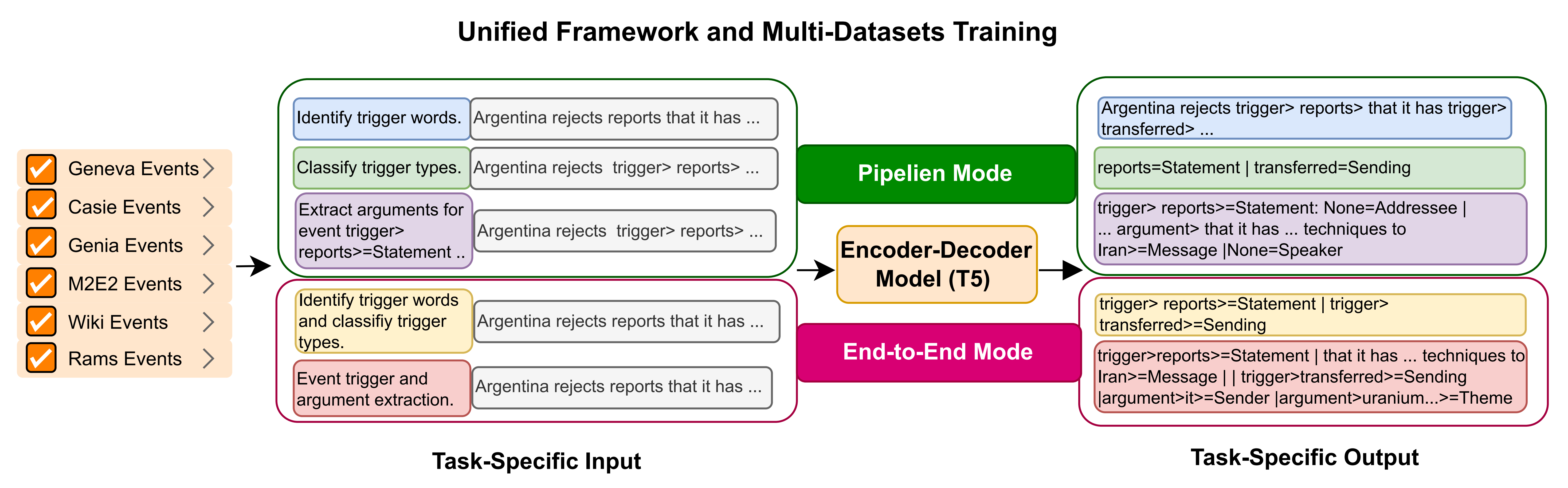}
    \caption{We train a unified sequence-to-sequence model based on T5 across six diverse domains, using all available datasets. Each dataset is mapped to a task-specific prompt and a corresponding output pattern to standardize event extraction. This formulation allows a single model to learn cross-domain event representations and supports both pipeline and E2E modes. }
    \label{fig:training_ee}
\end{figure*}

 In this work, we present an event extraction system based on a cross-domain unified generative framework using T5 architecture \cite{raffel2020exploring}. The framework formulates event extraction as a sequence-to-sequence task, enabling a single model to handle ED and EAE in a unified manner. Furthermore, the framework flexibly supports both pipeline and E2E modes. Based on this framework, we train a cross-domain backbone model on six  datasets spanning multiple domains jointly, enabling the model to learn representations that generalize across domains while retaining domain-specific semantics. We present a unified, cross-domain EE system implemented as a web-based application designed for practitioners to demonstrate the effectiveness and efficiency of the model trained jointly across diverse EE datasets.  






\section{Technology and Innovation}
We present a cross-domain unified generative framework for event extraction, designed to balance scalability, flexibility, and domain adaptation. The training framework is displayed in Figure~\ref{fig:training_ee} and the main technical contributions of our system are summarized as follows.

\begin{figure}[btp]
  \centering
    \resizebox{\linewidth}{!}{%
      \includegraphics{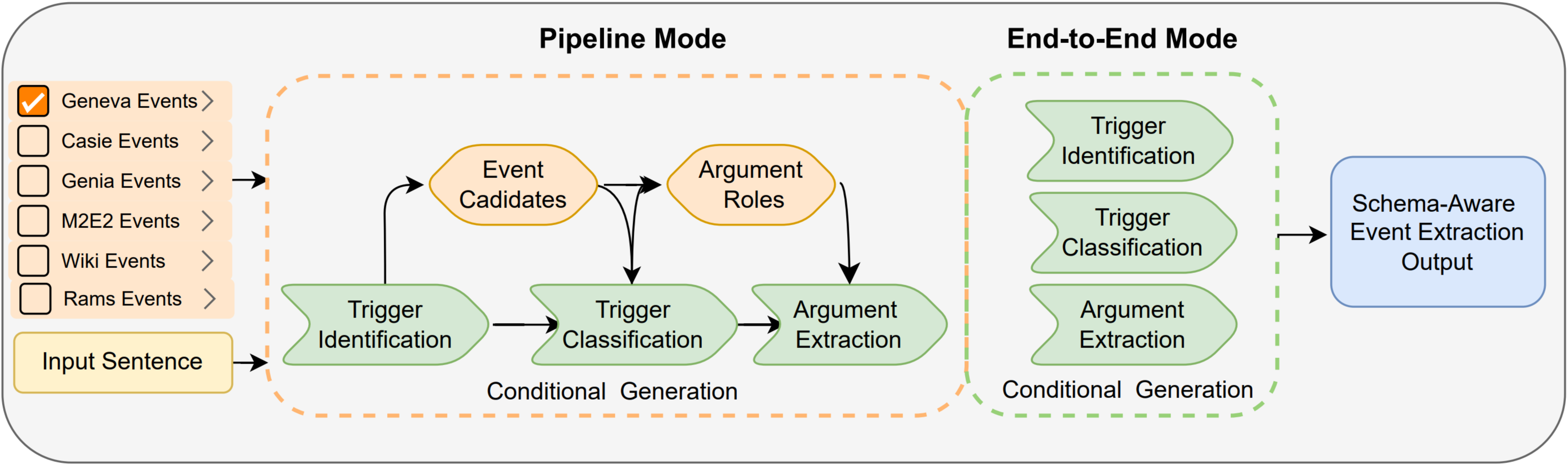}
    }
\caption{Pipeline and E2E event extraction workflows in inference time. }
    \label{fig:infer_modes}
  
\end{figure}

\paragraph{Unified generative modeling of event extraction tasks.} 
   Our framework formulates all event extraction tasks as a unified sequence-to-sequence task, handling trigger identification, event type classification and argument extraction simultaneously. This approach makes the system flexible, as it defines specific task codes and can handle changes to the input text without requiring modifications to the core architecture. Unifying modeling reduces the need for separate models, allowing consistent training across datasets while maintaining competitive performance in low-resource scenarios. 
\paragraph{Supporting both pipeline and E2E modes.} 
Our generative framework supports E2E and pipeline configurations in inference time, see Figure~\ref{fig:infer_modes}. Switching between modes requires using distinct input and target patterns for each mode for the same tasks as displayed in Figure~\ref{fig:training_ee}. The pipeline design provides explicit control over each extraction stage, supporting interpretability and stage-specific evaluation in inference time. In pipeline mode, the model detects event triggers in the input text. Each detected trigger is then classified to predict its event type based on the selected dataset schema, and finally, argument extraction is performed based on the predicted event type and its associated argument roles from the selected dataset schema. By contrast, E2E jointly performs trigger identification and event type classification in a single prediction step, simplifying inference and reducing error propagation, albeit with less fine-grained control than the pipeline workflow. A fully E2E formulation that incorporates argument extraction is also feasible, though more challenging due to the increased complexity of the output space.

\paragraph{Handling cross-domain and evolving label spaces.} To further enhance the generative framework, we extend the task-only conditioning paradigm by adding both a task identifier and a domain identifier to the beginning of the input sequence. This yields the unified input format: \texttt{\{\textit{dataset\_name}\}+\{\textit{task\_prefix}\}+\{\textit{text}\}}. The dataset name serves as a domain indicator and schema guide. This design allows the model to be trained jointly on multiple event extraction datasets, learning patterns that generalize across domains and accommodating diverse datasets and evolving schemas with minimal additional engineering while preserving domain-specific semantics. The six datasets used in our system follow the preprocessing of Huang~\shortcite{huang2024textee} and are summarized in Table~\ref{tab:dataset}. After fine-tuning, the cross-domain model can handle a wide range of event types simply by specifying the corresponding domain name in the input sequence, enabling the model predicting to the relevant domain.


\begin{table}[b]
\centering

\setlength{\tabcolsep}{3pt}
\small
\begin{tabularx}{\columnwidth}{|p{1.45cm}|p{1.25cm}|X|}
\hline
\textbf{Dataset} &
\textbf{\#Types} &
\textbf{Core Argument Roles} \\
\hline
CASIE (\textit{Cyber}) &
5 &
\textit{Attacker}, \textit{Target}, \textit{Malware}, \textit{Vulnerability}, \textit{Compromised Data}, \textit{Instrument} \\
\hline
GENEVA (\textit{Generic}) &
115 &
\textit{Protester}, \textit{Target}, \textit{Place}, \textit{Time}, \textit{Reason}, \textit{Organizer} \\
\hline
GENIA 2013 (\textit{Bio}) &
13 &
\textit{Theme}, \textit{Agent}, \textit{Site}, \textit{Resulting Entity} \\
\hline
M2E2 (\textit{Geo}) &
8 &
\textit{Location}, \textit{Victims}, \textit{Casualties}, \textit{Time} \\
\hline
RAMS (\textit{Newswire}) &
139 &
\textit{Participant}, \textit{Location}, \textit{Time}, \textit{Topic}, \textit{Outcome} \\
\hline
WikiEvents (\textit{Wiki}) &
50 &
\textit{Entity}, \textit{Organization}, \textit{Position}, \textit{Place}, \textit{Time}, \textit{Reason} \\
\hline
\end{tabularx}

\caption{Representative event types and their argument roles.}
\label{tab:dataset}
\end{table}

\section{Application Features}
\begin{figure}[h!]
    \centering
    \includegraphics[width=\linewidth]{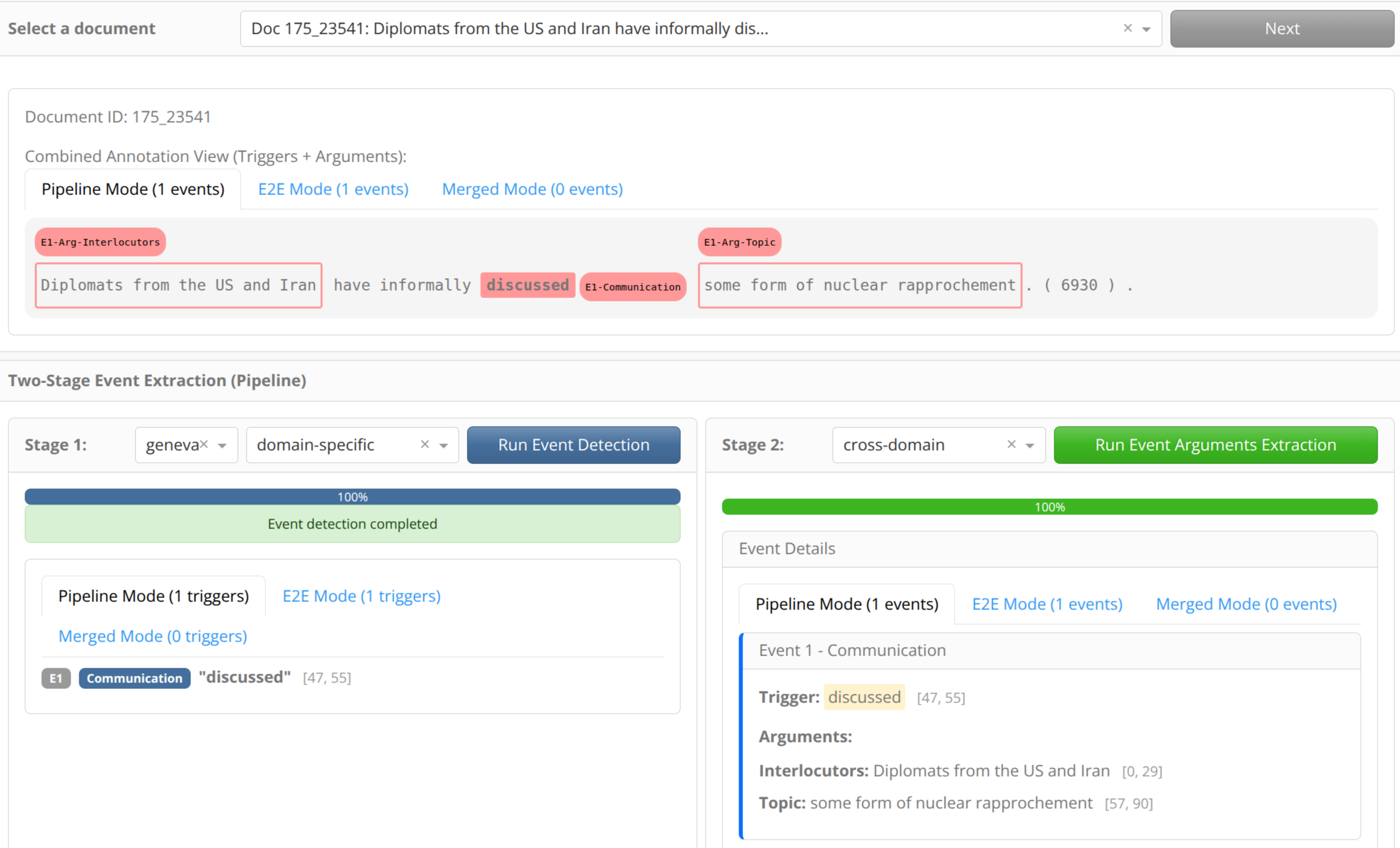}
   \caption{Two-stage event extraction interface. Stage 1 performs event detection using pipeline and E2E modes and merges the detected triggers. Stage 2 extracts arguments conditioned on the detected events and event-type-specific argument roles. Text highlighting supports result inspection.}
    \label{fig:two-stage}
\end{figure}

We present a web-based event extraction application based on the cross-domain unified generative framework. It allows JSON uploads and document selection with previews. Currently, it supports six event schemas (refer to Table~\ref{tab:dataset}) and provides domain-specific models trained on a single dataset and optimized for their respective domains, as well as a cross-domain model trained on all available datasets to generalize across domains. All models are fine-tuned on T5-base pre-trained model with approximately 220 million parameters. The application's main features are as follows.
\paragraph{Two-stage Event Extraction.}
Our event extraction system uses a two-stage workflow as displayed in Figure~\ref{fig:two-stage}. First, select the dataset (event schema) to define labels and roles, e.g., \textit{geneva}. The model performs event detection in two modes: pipeline and E2E. In pipeline mode, trigger spans are first identified and then classified into event types. In E2E mode, both steps are performed jointly in one pass. The outputs of the two modes are then merged into a consolidated trigger set in the first stage. Stage 2 then extracts arguments based on the event detection results of each mode obtained in the first stage. Model selection is configurable in both stages. This flexibility is motivated by practical observations that domain-specific models often achieve stronger trigger identification performance in high-resource domains, while cross-domain models tend to generalize better in low-resource settings, helping to mitigate error propagation in pipeline-based extraction. 
\begin{figure}[h!]
    \centering
    \includegraphics[width=\linewidth]{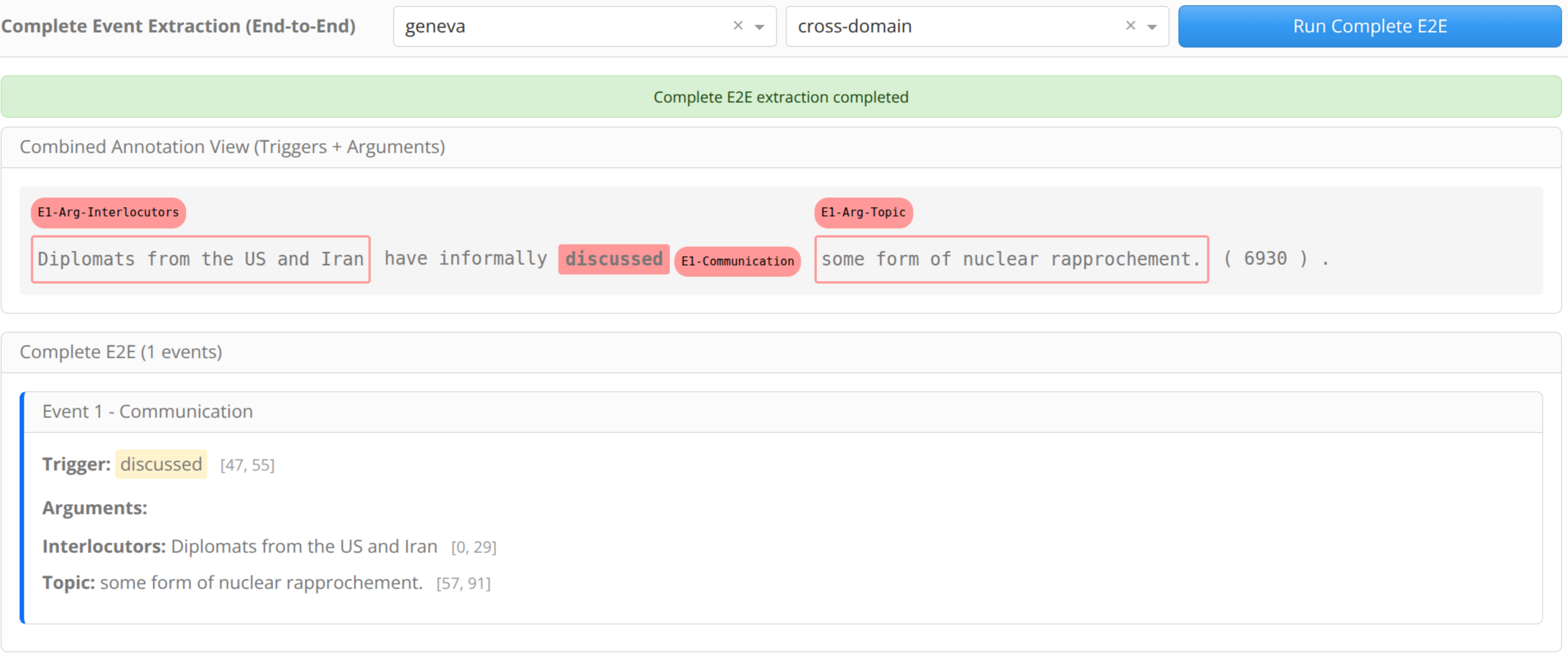}
    \caption{View of complete E2E event extraction, where the model runs all event extraction subtasks in one step. }
    \label{fig:e2e}
\end{figure}

\begin{figure}[h]
    \centering
    \includegraphics[width=\linewidth]{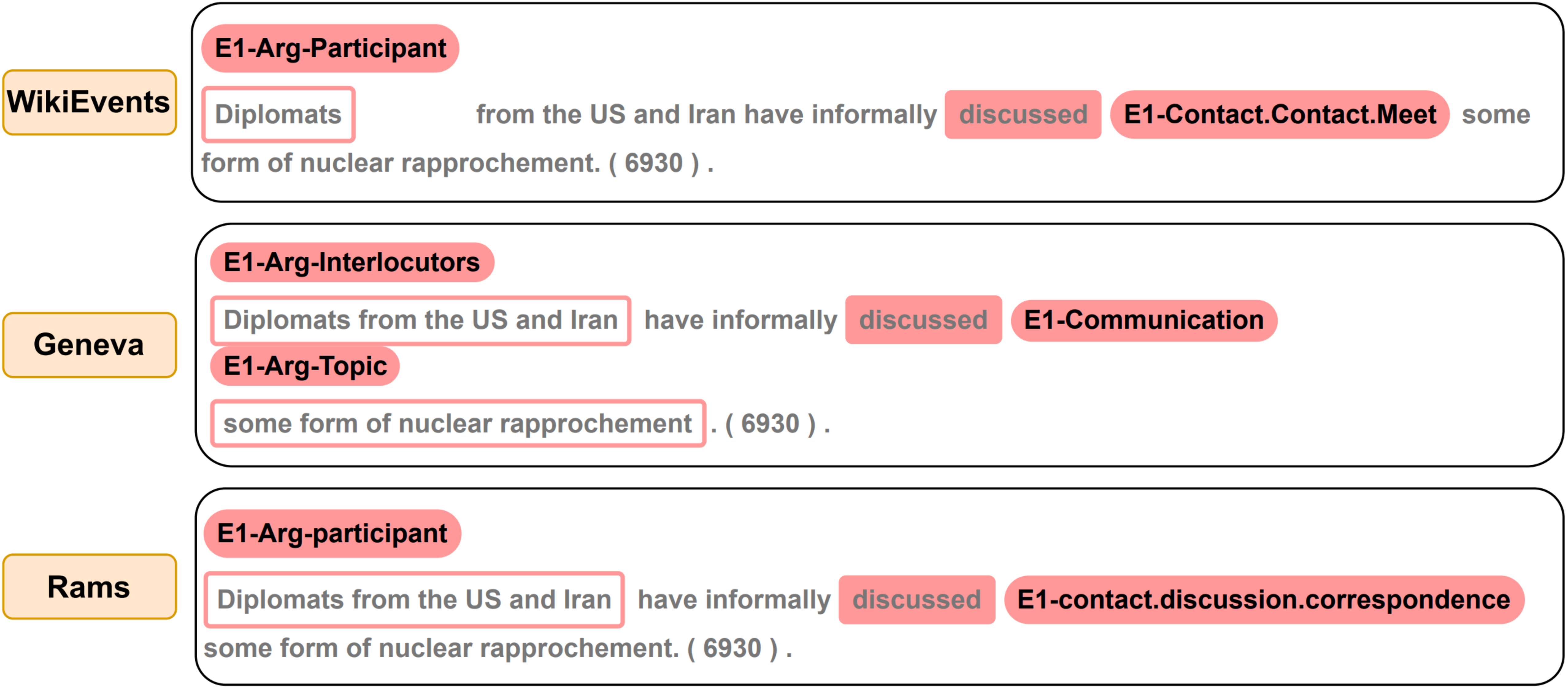}
    \caption{Event extraction on the same document with different dataset schemas.}
    \label{fig:multi-dataset-annotation}
\end{figure}

\paragraph{Results visualization comparison for all configurations.} In two-stage extraction mode, the visualization design is central: precise offsets, aligned highlights, and event cards make cross-mode differences explicit, so users can scrutinize Stage 1 outputs, decide which detection mode to trust, and see how that choice propagates into Stage 2 arguments. Beyond the two-stage workflow, the system supports full E2E extraction in a single step for efficient large-scale processing, and the interface presents all modes (\ref{fig:e2e}) in a consistent, comparable layout. Results can be examined per document and saved for side-by-side comparison, enabling informed configuration selection for each document.
\paragraph{Schema-aware multi-dataset annotation.}
The backend cross-domain fine-tuned model allows applying multiple event schemas to the same document while keeping each dataset's labels and structure separate, see Figure~\ref{fig:multi-dataset-annotation}. All event extraction results are saved with schema-aware labels that preserve dataset provenance across stage 1/2 and complete E2E outputs. The interface enables repeated annotation of the same document under different event schemas, producing a hierarchical, reusable JSON artifact suitable for downstream analysis and comparison.

\section{Conclusion}
In this work, we propose a unified generative approach for cross-domain event extraction. We demonstrate the system across multiple datasets spanning diverse domains and show robust performance across datasets, extraction configurations, and ontology sizes. We release an open-source application and video at: \href{https://youtu.be/8hNb-fAtcEA}{Youtube} and \href{https://github.com/sitingGZ/cross-domain-event-extraction-demo}{GitHub} that deliver a complete workflow from document ingestion to structured event outputs, supports both pipeline and E2E extraction, and exposes domain-specific and cross-domain models. The design prioritizes practical deployment, flexible adaptation to large label spaces, and reproducible analysis without extensive human redesign or dataset-specific customization, making it a scalable solution for real-world event extraction applications.

\clearpage

\section*{Ethical statement}
During the preparation of this work, we used GitHub Copilot to assist with the development of the code. After using this AI tool, we carefully reviewed and validated all the generated code. We take full responsibility for the accuracy, integrity and final content of this publication.

\bibliographystyle{named}
\bibliography{ijcai26}

\end{document}